\documentclass[runningheads]{llncs}
\usepackage{graphicx}
\usepackage[utf8]{inputenc}
\usepackage{epsfig}
\usepackage{calc}
\usepackage{amssymb}
\usepackage{amstext}
\usepackage{amsmath}
\usepackage{multicol}
\usepackage{pslatex}
\usepackage{amsmath}
\usepackage{listings}
\usepackage[ruled,vlined,linesnumbered]{algorithm2e}
\usepackage{comment}
\usepackage{amssymb}
\usepackage{url}
\usepackage{enumitem} 
\usepackage{mathbbol}  
\usepackage{hyperref}

\begin{document}

\title{At-Most-One Constraints in\\Efficient Representations of Mutex Networks}
\titlerunning{AMO Constraints in Efficient Representations of Mutex Networks}

\author{Pavel Surynek\orcidID{0000-0001-7200-0542}}
%\author{Paper Number: 77}
%
\authorrunning{P. Surynek}
%\authorrunning{Paper Number: 77}

\institute{
Faculty of Information Technology\\
Czech Technical University in Prague\\
Th\'{a}kurova 9, 160 00 Praha 6, Czechia\\
\email{pavel.surynek@fit.cvut.cz}
}

%\institute{
%\phantom{.}
%\\ \phantom{.}
%\\ \phantom{.}
%\email{\phantom{.}}
%}

\maketitle

\begin{abstract}
The At-Most-One (AMO) constraint is a special case of cardinality constraint that requires at most one variable from a set of Boolean variables to be set to $\mathit{TRUE}$. AMO is important for modeling problems as Boolean satisfiability (SAT) from domains where decision variables represent spatial or temporal placements of some objects that cannot share the same spatial or temporal slot. The AMO constraint can be used for more efficient representation and problem solving in mutex networks consisting of pair-wise mutual exclusions forbidding pairs of Boolean variable to be simultaneously $\mathit{TRUE}$. An on-line method for automated detection of cliques for efficient representation of incremental mutex networks where new mutexes arrive using AMOs is presented. A comparison of SAT-based problem solving in mutex networks represented by AMO constraints using various encodings is shown.

\end{abstract}

\section{Introduction}

We address the problem of representing the incremental mutex network efficiently using the At-Most-One (AMO) constraint \cite{DBLP:conf/soict/NguyenM15,DBLP:conf/cp/SilvaL07,DBLP:conf/cp/BailleuxB03,DBLP:conf/cp/Sinz05,DBLP:conf/isaim/BarahonaHN14}. The problem is addressed in the context of Boolean satisfiability (SAT) \cite{DBLP:conf/stoc/Cook71,DBLP:journals/ai/GentW96a,DBLP:journals/corr/abs-1910-00128,DBLP:conf/cp/Walsh00,Biere:2009:HSV:1550723} where the task is to decide whether there exists a truth value assignment satisfying a given Boolean formula. Usually we assume that formula $\mathcal{F}$ is specified using the {\em conjunctive normal form} (CNF) \cite{10030021172} as a finite list of {\em clauses} where each clause is a (finite) disjunction of {\em literals} and literal is either a variable or a {\em negation} of a variable. Let us denote a set of Boolean decision variables on top of which  $\mathcal{F}$ is expressed $Var(\mathcal{F})=X=\{x_1,x_2,...,x_n\}$.

A mutex network over set of variables $X$ is defined as a set of pair wise mutual exclusions that forbid a pair of propositional variables to be simultaneously $\mathit{TRUE}$. Formally the mutex network of size $k \in \mathbb{N}$ denoted $\mathcal{M}_k$ is a set of clauses $\big\{(\neg x_{u_i} \vee \neg x_{v_i}) \;|\; u_i,v_i \in \{1,2,...,n\}  \;\; \forall i = 1,2,...,k \big\}$. Mutex network $\mathcal{M}_k$ can be regarded as a subset of clauses of given Boolean formula $\mathcal{F}$ (in such case all binary clauses of the formula are included in $\mathcal{M}_k$).

Mutex networks appear in many problems expressed through the means of SAT. Many difficult problems that arise in {\em circuit design} \cite{DBLP:journals/tcad/AloulRMS03}, {\em scheduling} \cite{DBLP:journals/jcp/AloulZAAA13}, {\em classical planning} \cite{DBLP:conf/ijcai/KautzS99,DBLP:conf/aaai/KautzS96,DBLP:conf/ecai/Rintanen12a,DBLP:conf/socs/FroleyksBS19} or various cases of {\em domain dependent planning} \cite{DBLP:conf/socs/Surynek12,DBLP:conf/aips/PandeyR18} can be expressed using networks of mutual exclusions. Generally mutex networks are often a product of spatial and temporal constraints between some objects. In real physical domains, objects usually cannot share the same spatial or temporal slot which directly leads to mutual exclusions of occurence of different objects in the same slot or the same object in multiple slots. Using common {\em direct encodings} \cite{DBLP:books/sp/Petke15} is which decision Boolean variables are used directly to represent some object relations (for example $x_1^A$ and $x_1^B$ represent occurence of object $A$ and object $B$ in slot $1$ respectively) can often give rise to complex mutex network (consisting of clauses like $\neg x_1^A \vee \neg x_1^B$ saying that objects $A$ and $B$ cannot be simultaneously in slot 1).

\subsection{Contribution}

We contribute by a method for automated detection of the AMO constraint in mutex networks. We propose algorithms for detecting cliques in mutex networks that can operate in an on-line mode which means that mutexes are processed as they arrive into the mutex network. Such feature is of great interest in domains where lazy Boolean encodings are used. Several Boolean encodings of the AMO constraint are recalled. After detecting cliques in mutex network these cliques can be substituted by AMO constraint using the encoding of user's choice.

The paper is organized as follows: first we introduce existing encodings of the AMO constraint. Then an on-line clique detection algorithms are introduced: one exact and one relaxed. Both algorithms are theoretically analyzed. Finally, we implement and evaluate the suggested AMO substitution method on a set of benchmarks including various difficult SAT instances.

\section{Background}

The {\em At-Most-One} constraint (AMO) over Boolean variables $X_{\leq} = \{x_{i_j}\}_{j=1}^m$ with $i_j \in \{1,2,...,n\}$ for $j=1,2,...,m$ denoted $_{\leq 1}\{x_{i_1},x_{i_2},...x_{i_m}\}$ or $_{\leq 1}\{x_{i_j}\}_{j=1}^m$ requires that at most one Boolean variable from $x_{i_j}$ variables can be set to $\mathit{TRUE}$. AMO is a special case of more general {\em cardinality constraints} \cite{DBLP:conf/cp/BailleuxB03,DBLP:conf/cp/Sinz05} denoted $_{\leq c}\{x_{i_1},x_{i_2},...x_{i_m}\}$ with $c \in \mathbb{N}$ requiring that at most $c$ variables from $X_{\leq}$ are assigned $\mathit{TRUE}$.

Cardinality constraints have wide use in problem modeling since they enable bounding various numeric values in inside the yes/no environment of Boolean formulae. This is a significant contribution that brings SAT towards practical use in problem solving \cite{4605925}.

\begin{definition} {\bf (clique clustering).} Given an undirected graph $G=(V,E)$, a {\em clique clustering} of $G$ is a collection of subsets of vertices, $\mathcal{C}=\{K \subseteq V\}$ such that each $C \in \mathcal{K}$ induces a complete sub-graph of $G$ and each edge of $G$ is covered by $\mathcal{C}$, that is, $\forall \{u,v\} \in E(\exists K \in \mathcal{C} \text{ such that } u,v \in K$).
\end{definition}

The concept of clique clustering can be naturally converted for mutex networks by substituting of indices Boolean variables from $X$ instead of $V$ and the set of binary clauses of $\mathcal{M}_k$ instead of $E$ (that is, clause $(\neg x_{u_i} \vee \neg x_{v_i})$ represents edge $\{u_i,v_i\}).$

\subsection{Encodings of the At-Most-One Constraint}

The At-Most-One constraint can be expressed by various encodings often using additional auxiliary Boolean variables. The set of auxiliary variables is denoted  $A_{\leq}$ in this context. The basic encoding of AMO often called {\em pairwise} can simply use the mutex network of size $\frac{m \cdot (m - 1)}{2}$ consisting of clauses $\big\{\neg x_{u_i} \vee \neg x_{v_i} \;|\; u_i,v_i \in \{i_1,i_2,...,$ $i_m\} \wedge u_i < v_i \big\}$. Even though this encodings supports achieving {\em arc-consistency} \cite{DBLP:journals/ai/Mackworth77} through {\em unit propagation} \cite{journals/jlp/DowlingG84,DBLP:conf/ecai/Gent02} it lacks any understanding of the global structure of the constraint \cite{DBLP:conf/sara/Surynek07}. It has been experimentally shown that different encodings of the At-Most-One constraint significantly outperform the pairwise encoding in a number of benchmarks \cite{DBLP:conf/soict/NguyenM15}.

\subsubsection{Binary Encoding}

The {\em binary encoding} \cite{DBLP:journals/jar/FrischPDN05} uses the idea of mapping $m$ possible settings of exactly one variable in set $X_{\leq}$ to $\mathit{TRUE}$ to $m$ possible different settings of bits in a bit vector of length $\lceil{\log_2{m}}\rceil$. The binary encoding uses $\lceil{\log_2{m}}\rceil$ auxiliary variables:  $A_{\leq} = \{b_1, b_2, ..., b_{\lceil{\log_2{m}}\rceil}\}$ to represent the bit vector. Whenever $x_{i_j}$ is set to $\mathit{TRUE}$ the bit vector variables must be set following the binary encoding to represent value $j$ in the bit vector. Let $\mathbb{b}_l^j=
\begin{cases}
b_l & \text{if } l\text{-th bit of binary encoding of }j \text{ is } 1 \\
\neg b_l & \text{if } l\text{-th bit of binary encoding of }j \text{ is } 0
\end{cases}$

Specifically we have the following formula to represent $_{\leq 1}\{x_{i_j}\}_{j=1}^m$ using the binary encoding:

\begin{gather*}
(\neg x_{i_1} \vee \mathbb{b}_1^1) \wedge (\neg x_{i_1} \vee \mathbb{b}_2^1) \wedge ... \wedge (\neg x_{i_1} \vee \mathbb{b}_{\lceil{\log_2{m}}\rceil}^1) \wedge \\
(\neg x_{i_2} \vee \mathbb{b}_1^2) \wedge (\neg x_{i_2} \vee \mathbb{b}_2^2) \wedge ... \wedge (\neg x_{i_2} \vee \mathbb{b}_{\lceil{\log_2{m}}\rceil}^2) \wedge \\
...\\
(\neg x_{i_m} \vee \mathbb{b}_1^m) \wedge (\neg x_{i_m} \vee \mathbb{b}_2^m) \wedge ... \wedge (\neg x_{i_m} \vee \mathbb{b}_{\lceil{\log_2{m}}\rceil}^m)
\end{gather*}

\subsubsection{Sequential Counter Encoding}
The {\em sequential encoding} \cite{DBLP:conf/cp/Sinz05} simulates calculation of the number of Boolean variables set to $\mathit{TRUE}$ by the circuit for calculating sums. The calculation uses unary encoding of the partial sum. Intuitively, the calculation can be regarded as a loop going across  $X_{\leq}$. Each time a positively assigned variable in $X_{\leq}$ is encountered, the partial sum is increased by one. This approach nicely works for general cardinality constraint  $_{\leq c}\{x_{i_1},x_{i_2},...x_{i_m}\}$. Enforcing the cardinality constarint can done by bounding the partial sum by $c$. In the AMO case, the situation is much simpler. It is sufficient to bound the partial sum by $1$.

Technically we need to introduce an array of auxiliary variables $S=\{s_{l,j}\;|\;l=1,2,...m+1; j = 1,2\}$ representing partial sums for individual steps of the main loop using the unary calculation. The interpretation is that $s_{l,j}$ is $\mathit{TRUE}$ if the sum of variables set to $\mathit{TRUE}$ among $\{x_{i_1},x_{i_2},...,x_{i_l-1}\}$ is $j$. The following clauses carry out the partial sums:

\begin{gather*}
(\neg x_{i_1} \vee \neg s_{1,1} \wedge s_{2,2}) \wedge
(\neg x_{i_1} \vee s_{1,1}) \wedge
(x_{i_1} \vee \neg s_{1,1} \wedge s_{2,1}) \wedge
(x_{i_1} \vee \neg s_{1,2} \wedge s_{2,2}) \wedge \\
(\neg x_{i_2} \vee \neg s_{2,1} \wedge s_{3,2}) \wedge
(\neg x_{i_2} \vee s_{2,1}) \wedge
(x_{i_2} \vee \neg s_{2,1} \wedge s_{3,1}) \wedge
(x_{i_2} \vee \neg s_{2,2} \wedge s_{3,2}) \wedge \\
...\\
(\neg x_{i_m} \vee \neg s_{m,1} \wedge s_{m+1,2}) \wedge
(\neg x_{i_m} \vee s_{m,1}) \wedge
(x_{i_m} \vee \neg s_{m,1} \wedge s_{m+1,1}) \wedge
(x_{i_m} \vee \neg s_{m,2} \wedge s_{m+1,2})
\end{gather*}

Finally, we need to enforce $s_{m+1,2}$ to $\mathit{FALSE}$ by introducing unit clause $\neg s_{m+1,2}$ to bound the last partial sum to at most one.

\subsubsection{Product Encoding}

The {\em product encoding} \cite{Chen2010ANS} relies on the idea of assigning each variable a point in a 2-dimensional array of Boolean variables and projecting this array to one dimension. In both one dimensional projection it is required that at most one variable is set to $\mathit{TRUE}$ which implies that there is at most one in original variables set to $\mathit{TRUE}$. In other words, if say two (or more) distinct variables from $X_{\leq}$ are set to $\mathit{TRUE}$ then at least in one of the two projections we can see two $\mathit{TRUE}$ values because the two variables must differ in at least one of their projections. Having two (or more) $\mathit{TRUE}$ values in any of the projections is however forbidden by the encoding.

More precisely we use the set of auxiliary variables $A_{\leq} = \{p_1, p_2, ..., p_d, q_1, q_2, ...,$ $q_d\}$ representing the 2D array of Booleans where $d = \lceil{\sqrt{m}}\rceil$ represents the size of projecting dimension. Each point from the 2D array is assigned to a distinct variable from $X_{\leq}$, that is, following clauses are introduced in the encoding:

\begin{gather*}
(\neg x_{i_1} \vee p_1) \wedge (\neg x_{i_1} \vee q_1) \wedge (\neg x_{i_2} \vee p_1) \wedge (\neg x_{i_2} \vee q_2) \wedge \\ ... \wedge (\neg x_{i_d} \vee p_1) \wedge (\neg x_{i_d} \vee q_d) \wedge \\
(\neg x_{i_{d+1}} \vee p_2) \wedge (\neg x_{i_{d+1}} \vee q_1) \wedge (\neg x_{i_{d+2}} \vee p_2) \wedge (\neg x_{i_{d+2}} \vee q_2) \wedge \\ ... \wedge (\neg x_{i_{2d}} \vee p_2) \wedge (\neg x_{i_{2d}} \vee q_d) \wedge \\
... \\
(\neg x_{i_{(d-1)d+1}} \vee p_d) \wedge (\neg x_{i_{(d-1)d+1}} \vee q_1) \wedge (\neg x_{i_{(d-1)d+2}} \vee p_d) \wedge (\neg x_{i_{d-1)d+2}} \vee q_2) \wedge \\ ... \wedge (\neg x_{i_{m}} \vee p_d) \wedge (\neg x_{i_{m}} \vee q_{m-(d-1)d})
\end{gather*}

Let us note that if $m$ is not a square of some $d \in \mathbb{N}$ then not all points in the 2D Boolean array has assigned a variable from $X_{\leq}$. Also we note that is not required that both projection dimensions are of the same size. A 2D array with different sizes of both dimensions can be used as well:  $A_{\leq} = \{p_1, p_2, ..., p_{d_1}, q_1,$ $q_2, ..., q_{d_2}\}$ such that $d_1 \cdot d_2 \geq m$.

In addition to above clauses it needs to be ensured that at most one variable in each projection is set to $\mathit{TRUE}$ which can be done by introducing following AMO constraints: $_{\leq 1}\{p_1,p_2,...p_d\}$ and $_{\leq 1}\{q_1,q_2,...q_d\}$ that are of smaller size than the originally encoded AMO so inductively any encoding can be used for them the product encoding including.

\subsubsection{Commander Encoding} The {\em commander encoding} \cite{Klieber2007EfficientCE} partitions $X_{\leq}$ into disjoint subsets $Y_1=\{y_1^1,...y_{g_1}^1\}$, $Y_2=\{y_1^2,...,y_{g_2}^2\}$, ... $Y_d=\{y_1^d,...,y_{g_d}^d\}$. For each group of variables $Y_i$ we introduce an auxiliary commander variable $c_i$. The interpretation is that $c_i$ set to $\mathit{TRUE}$ selects a candidate from group $Y_i$. In other words, commander variables introduce a hierarchical structure for selecting at most one variable from the original set $X_{\leq}$: at most one commander variable can be set to $\mathit{TRUE}$ while it permits to select at most one variable from its group; no other variable in other groups can be set to $\mathit{TRUE}$. Following clauses need to be introduces to carry out the encoding:

\begin{gather*}
(\neg c_1 \vee y_1^1 \vee y_2^1 \vee ... y_{g_1}^1) \wedge (\neg c_2 \vee y_1^2 \vee y_2^2 \vee ... y_{g_2}^2) \wedge ... \wedge (\neg c_d \vee y_1^d \vee y_2^d \vee ... y_{g_d}^d)
\end{gather*}

In addition to this following AMO constraints need to be introduced: $_{\leq 1}\{\neg c_1 \vee y_1^1 \vee y_2^1 \vee ... y_{g_1}^1\}$,  $_{\leq 1}\{\neg c_2 \vee y_1^2 \vee y_2^2 \vee ... y_{g_2}^2\}$, ...,  $_{\leq 1}\{\neg c_d \vee y_1^d \vee y_2^d \vee ... y_{g_d}^d\}$. Finally, we need to enforce that at most one commander variable is selected by: $_{\leq 1}\{c_1 \vee c_2 \vee ... c_{d}\}$.

The number of groups partitioning $X_{\leq}$ is set to $d = \lceil{\sqrt{m}}\rceil$ which is a good compromise between the number of groups and their size.

\section{On-Line Clique Detection in Mutex Networks}

We present here techniques for detecting {\em cliques} (complete sub-graph) in mutex networks. Having cliques detected on top of set of mutexes one can introduce AMO constraints using some more advanced encoding instead the set of mutex clauses. Finding cliques in an undirected graph is well known NP-hard problem hence solving it exactly is as hard as solving the original formula satisfiability problem. Therefore certain trade-offs between the quality of cliques being discovered and complexity of the method must be adopted.

We first present an exact exponential time/space algorithm for finding all possibly overlapping cliques in the input mutex network. Then we will show how to relax the algorithm to reduce its complexity to acceptable polynomial level while keeping its original idea.

\subsection{Exact Algorithm}

The exact algorithm for clique detection in mutex network is presented using pseudo-code as Algorithm \ref{alg-Exact-Clique}. The algorithm relies on the idea of merging variables into clusters while valued meta-edges between clusters are maintained. The meta-edge between clusters corresponds to the set of edges interconnecting the two clusters. The value of meta-edge is the number of edges connecting the clusters.

When new mutexes arrive to the mutex network the algorithm updates (increments) the value of meta-edges between all pairs of clusters the mutex variables participate in (line 8 in \textsc{Exact-Cliques}). If this value of the meta-edge reaches the size of the complete graph between the pair of cluster, the clusters are merged together to form a new cluster (the original pair of clusters is maintained). This is done in the \textsc{Increment-Cluster-Link} procedure.

\begin{algorithm}[t]
\begin{footnotesize}
\SetKwBlock{NRICL}{\textsc{Exact-Cliques}($\mathcal{M}_k = \big\{(\neg x_{u_i} \vee \neg x_{v_i}) \;|\; u_i,v_i \in \{1,2,...,n\} \; \forall k = 1,2,...,k\big\}$)}{end} \NRICL{
    let $\mathcal{E}(K_v,K_u) = 0$ for any $K_u,K_v \subseteq \{1,2,...,n\}$\\
    
    \For {each $i=1,2,...,n$}{
        $\mathcal{C} \gets \{\{i\}\}$
    }

    \For {each $(\neg x_{u_i} \vee \neg x_{v_i}) \in \mathcal{M}_k$}{
    	  \For {each $K_u \in \mathcal{C}$ such that $u_i \in K_u$}{
    	      \For {each $K_v \in \mathcal{C}$ such that $v_i \in K_v$}{
    	      	   \textsc{Increment-Cluster-Link}($K_u$, $K_v$)
    	      }    	  
    	  }
    }
    \Return $\mathcal{C}$
}

\SetKwBlock{NRICL}{\textsc{Increment-Cluster-Link}($K_u$,$K_v$)}{end} \NRICL{
	$\mathcal{E}(K_u,K_v) \gets \mathcal{E}(K_u,K_v) + 1$\\
	
	\If{$\mathcal{E}(K_u,K_v) = |K_u \setminus (K_u \cap K_v)| \times |K_v \setminus (K_v \cap K_v)|$}{
	
	      \For{each $K \in \mathcal{C}$ such that $\mathcal{E}(K,K_u) > 0 \vee \mathcal{E}(K,K_v)>0$}{
		    $\mathcal{E}(K, K_u \cup K_v) \gets \mathcal{E}(K, K_u) + \mathcal{E}(K, K_v) - \mathcal{E}(K, K_u \cap K_v)$\\
		 }
		$\mathcal{C} \gets \mathcal{C} \cup \{K_u \cup K_v\}$		    
	}
}
\caption{Exact clique detection algorithm.} \label{alg-Exact-Clique}
\end{footnotesize}
\end{algorithm}

Without proof let us state that the \textsc{Exact-Cliques} algorithm is sound and complete and returns clique clustering of the input mutex network. All cliques contained in the input mutex network are detected by the algorithm, that is, if a subset of variables $X \subseteq X_{\leq}$ induces a clique of mutexes  w.r.t. $\mathcal{M}_k$ then $X \in \mathcal{C}$.

To provide deeper insight into how the algorithm proceeds we summarize its complexity in the following proposition.

\begin{proposition} {\bf (\textsc{Exact-Cliques} complexity).}
The \textsc{Exact-Cliques} algorithm for mutex network $\mathcal{M}_k$ over set of variables $X=\{x_1,x_2,...,x_n\}$ has the worst case time complexity $\mathcal{O}(k\cdot 2^{3n})$ and the worst case space complexity $\mathcal{O}(2^{2n})$.
\end{proposition}

\noindent
{\bf Proof.} Let us first calculate the complexity of the procedure for incrementing the number of links between a pair of clusters $K_u$ and $K_v$ (\textsc{Increment-Cluster-Link}). The determining factor is updating the number of links of the newly created cluster with existing clusters (lines 13-14). This consumes time of $\mathcal{O}(2^n)$ since there is up to $2^n$ clusters that need to update their number of links towards the new cluster.

The dominating factor in the overall space complexity is the structure for keeping the number of links between variable clusters represented by $\mathcal{E}$. As the number of clusters is at most $2^n$, the number of links connecting clusters is bounded by $2^{2n}$. Hence the overall space complexity of $\mathcal{O}(2^{2n})$.

To calculate time complexity we need to observe that single variable can participate in as many as $2^n$ clusters. Hence line 8 in \textsc{Exact-Cliques} where the number of links between clusters is incremented can be executed as many as $2^{2n}$ times per one mutex which gives altogether $k \cdot 2^{2n}$ execution across entire mutex network. Taking into account the $\mathcal{O}(2^n)$ time required by the single link increment we obtain that the algorithm needs time $\mathcal{O}(k\cdot 2^{3n})$ steps in the worst case. $_\blacksquare$\\

The algorithm of such high complexity is impractical however our preliminary experiments indicate that it can be well used for detecting cliques in small mutex networks.

Observe that the \textsc{Exact-Cliques} algorithm can operate in an {\em on-line} mode where new mutexes arrive piece by piece while the clique clustering is still kept up to date. This is important in many applications such as planning where the SAT model of a problem is often incrementally modified (that is new constraints including mutexes are added) so we do not need to search for cliques from scratch but only update recent changes

We will use the algorithm as a starting point for a relaxed version which will keep the on-line functionality.

\subsection{Relaxation of the Exact Algorithm}

We relax the exact clique detection algorithm while keeping its high level structure of merging the variable clusters. We do this by restricting the set of pair of clusters for which merging attempt is made. Intuitively, as we aim on finding large cliques, it seems to be promising to focus merging of large clusters together while smaller clusters are omitted.

The method presented using pseudo-code in Algorithm \ref{alg-Relaxed-Clique} attempts to merge only the largest pair of clusters. That is, for an arriving mutex $(\neg x_{u_i} \vee \neg x_{v_i})$ we identify largest cluster $K^*_u$ containing $u_i$ and similarly largest cluster $K^*_v$ containing $v_i$ (lines 6-7). The attempt to merge clusters is done only for $K^*_u$ and $K^*_v$.

\begin{algorithm}[t]
\begin{footnotesize}
\SetKwBlock{NRICL}{\textsc{Relaxed-Cliques}($\mathcal{M}_k = \big\{(\neg x_{u_i} \vee \neg x_{v_i}) \;|\; u_i,v_i \in \{1,2,...,n\} \; \forall k = 1,2,...,k\big\}$)}{end} \NRICL{
    let $\mathcal{E}(K_v,K_u) = 0$ for any $K_u,K_v \subseteq \{1,2,...,n\}$\\
    
    \For {each $i=1,2,...,n$}{
        $\mathcal{C} \gets \{\{i\}\}$
    }
    
    \For {each $(\neg x_{u_i} \vee \neg x_{v_i}) \in \mathcal{M}_k$}{
        $K^*_u \gets \text{argmax}_{K_u \in \mathcal{C} | u_i \in K_u} |K_u|$ \\
        $K^*_v \gets \text{argmax}_{K_v \in \mathcal{C} | v_i \in K_v} |K_v|$ \\           
        
%    	  \For {each $K_u \in \mathcal{C}$ such that $u_i \in K_u \wedge |K_u| = s_u$}{
%    	      \For {each $K_v \in \mathcal{C}$ such that $v_i \in K_v \wedge |K_v| = s_v$}{
   	  \textsc{Increment-Cluster-Link}($K^*_u$, $K^*_v$)\\
   	  $ \mathcal{C} \gets \mathcal{C} \cup \{\{u_i,v_i\}\}$
%    	      }    	  
%    	  }
    }
    \Return $\mathcal{C}$
}
\caption{Relaxed clique detection algorithm.} \label{alg-Relaxed-Clique}
\end{footnotesize}
\end{algorithm}

\begin{proposition} {\bf (\textsc{Relaxed-Cliques} complexity).}
The \textsc{Relaxed-Cliques} algorithm for mutex network $\mathcal{M}_k$ over set of variables $X=\{x_1,x_2,...,x_n\}$ has the worst case time complexity $\mathcal{O}(k^2)$ and the worst case space complexity $\mathcal{O}(k^2)$.
\end{proposition}

\noindent
{\bf Proof.} We first need to observe that single mutex being processed $\mathcal{M}_k$ in the main loop (lines 6-8) can give rise to at most new cluster of variables. Hence the total number of clusters is bounded by $k$. The remaining calculation of the complexity relies on this bound (we assume that $n < k$). Incrementing the number of links between a pair of clusters then takes $k$ steps since we need to update connection of the new cluster towards at most $k$ existing clusters (lines 13-14 of \textsc{Increment-Cluster-Link}).

For each of $k$ mutexes in input mutex network $\mathcal{M}_k$ we take the largest cluster in which its variables participate and increment the number of links between the pair of largest clusters. Assuming $\mathcal{O}(k)$ the worst case time complexity of incrementing the main loop of \textsc{Relaxed-Cliques} (lines 5-8) yields time $\mathcal{O}(k^2)$ altogether.

The worst case space complexity is determined by the need to represent number of links between clusters $\mathcal{E}$ which can be done using space $\mathcal{O}(k^2)$.  $_\blacksquare$\\

\begin{figure}[h]
    \centering
    \includegraphics[trim={1.0cm 20.5cm 2.0cm 2.5cm},clip,width=1.0\textwidth]{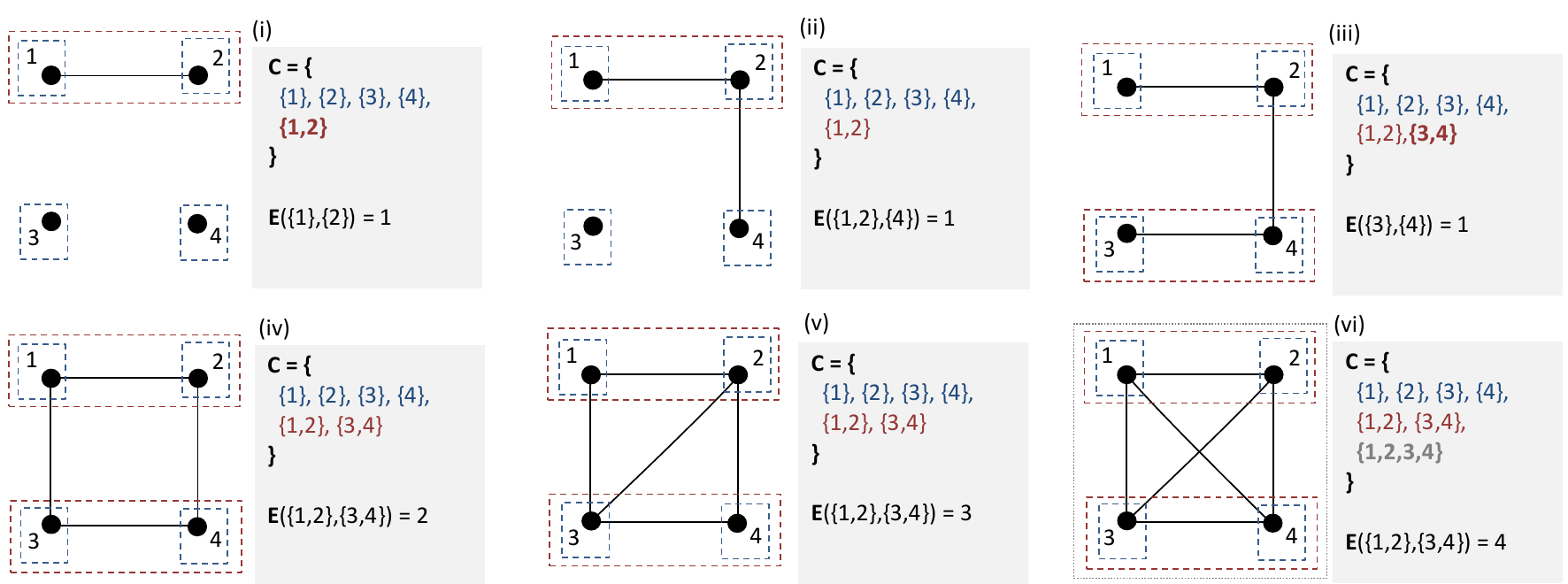}
    \caption{The process of clique clustering construction by the \textsc{Relaxed-Cliques} algorithm. Mutexes of a network consisting of single clique on 4 variables $\{x_1,x_2,x_3,x_4\}$ are processed following the ordering: $\{1,2\}, \{2,3\}, \{3,4\}, \{1,3\}, \{2,3\}$ and $\{1,4\}$. Eventually the original clique covering all 4 variables is detected in step (vi). }
    \label{fig-cluster-A}
\end{figure}

It is a question now if such dramatic reduction of time and space complexity through restricting the link incrementing only on largest clusters keeps ability of the algorithm to detect cliques reasonably. The example show in Figure \ref{fig-cluster-A} illustrates that the restriction on largest clusters does not compromise finding the clique.

An intuitive insight into how the relaxed clique clustering works suggests that it is important to add mutexes participating in a single clique in a sequence not much interrupted by additions of other mutexes. In such case, clusters have chance to grow to cover the clique and do not grow outside the clique. Practical applications suggests that this property is often the case since for example all objects related to given spatial slot that are spatially excluded are processed in a single uninterrupted block.

\subsection{The At-Most-One Constraint Substitution in Mutex Networks}
After detecting cliques in mutex network we can convert the encoding so that instead of the basic pair-wise representation of AMOs different encodings can be used. The soundness of substitution of AMO constraints relies on the property of clique clustering (proof omitted):

\begin{proposition}
Assume a clique clustering $\mathcal{C}$ of mutex network $\mathcal{M}_k$. Representing each clique $C \in \mathcal{C}$ using some encoding of the AMO constraint results in an equivalent instance as that represented by $\mathcal{M}_k$.
\end{proposition}

The equivalence here is defined as having identical set of satisfying assignments, that is the set of satisfying assignments of conjucntion of AMOs for cliques is the same as the set of satisfying assignment for the original set of mutexes $\mathcal{M}_k$ (auxiliary variables are ignored).

Moreover it is easy to see that if there are cliques $C' \in \mathcal{C} and C \in \mathcal{C}$ such that $C \subseteq C'$ then it is sufficient to represent $C'$ using the AMO constraint to keep the above equivalence valid.

\section{Experimental Evaluation}

We evaluated the suggested approach of AMO substitution in mutex networks experimentally. The approach can be used both in {\em eager} SAT encodings of problems where the target formula is first constructed in advance and then consulted with the SAT solver as well as in the {\em lazy setup} where the target formula is constructed incrementally piece by piece and the SAT solver is consulted multiple times during the process of construction \cite{DBLP:series/txtcs/KroeningS16}.

\subsection{Benchmarks and Setup}

The experimental evaluation is based on the \textsc{Glucose 3.0} SAT solver \cite{DBLP:conf/sat/AudemardLS13} which has been used as a library linked to the testing program. The test itself is implemented in C++ \footnote{To enable reproducibility of presented results we provide the source code and supporting experimental data on the author's website:\href{http://users.fit.cvut.cz/surynpav/research/mutexAMO2020}{http://users.fit.cvut.cz/surynpav/research/ mutexAMO2020}. The source code of presented algorithms is also available in author's Git repository: \href{http://github.com/surynek/mutEX}{http://github.com/surynek/mutEX}.}.

The AMO substituion is implemented for all discussed encodings: {\bf binary}, {\bf sequential}, {\bf product}, and {\bf commander} encodings while the {\bf pair-wise} is kept as the baseline encoding for reference comparison. We divided the experiments in three tests:

\begin{enumerate}[label=(\roman*)]
\item evaluation of AMO substitution in random mutex formulae and
\item evaluation of clique detection in random mutex network
\item evaluation of AMO substitution in standard SAT benchmarks consisting of difficult instances \cite{DBLP:journals/tcad/AloulRMS03}
\end{enumerate}

\subsection{Comparison of Mutex Network Representations Using AMOs}

In all tests we compared representation mutex network using detected AMO constraints and the base-line representation where mutexes are kept in their original form as pair-wise encoding. The test runs in three phases:

\begin{enumerate}
\item {\bf Clique clustering.} This phase processes the input SAT instance in CNF that is ether generated synthetically or read from the input file. All clauses from the input representing mutexes (clauses of the form $(\neg x \vee \neg y)$) are treated as being part of mutex network $\mathcal{M}_k$ and are not declared to the SAT solver; other clauses, that is those of higher arity than 2 and those not representing mutual exclusion of $\mathit{TRUE}$ assignment to a pair of variables are directly declared to the SAT solver. $\mathcal{M}_k$ consisting of collected mutexes is processed by the \textsc{Relaxed-Cliques} algorithm which produces clique clustering $\mathcal{C}$.

\item {\bf AMO encoding phase.} The clique clustering $\mathcal{C}$ is converted to actual AMO encoding so that the resulting formula is equivalent to the original one. We start from largest cliques in $\mathcal{C}$ and continue down to cliques of size 3. Each clique $C \in \mathcal{C}$ is checked if it is subsumed by any larger already processed clique. If not then $C$ is encoded using selected AMO encoding and resulting clauses are recorded. If $C$ is subsumed by some $C' \in \mathcal{C}$ then $C$ is simply ignored as its meaning is already captured by $C'$. The remaining clique of size 2 (simple mutual exclusions) are declared as clauses if they assumed they are not subsumed by any larger clique. 

\item {\bf SAT solving phase.} This phase corresponds to consulting the SAT solver with encoded instance. Encoded clauses are all declared to the SAT solver and the solver is started using its default setting.
\end{enumerate}

\begin{figure}[h]
    \centering
    \includegraphics[trim={2.0cm 17.0cm 3.0cm 2.5cm},clip,width=0.9\textwidth]{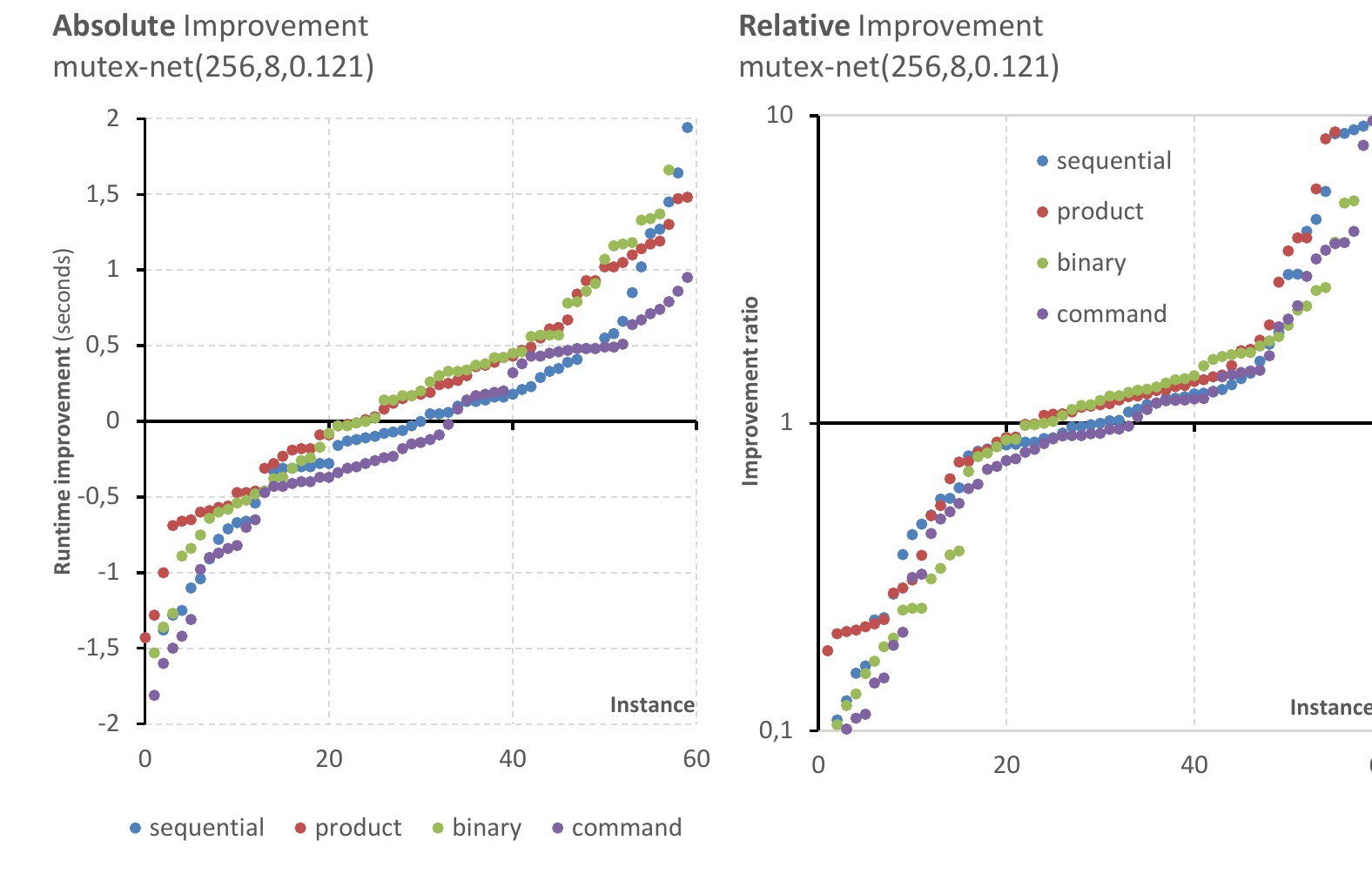}
    \caption{Absolute and relative improvement by AMOs using various encodings in random mutex network with $N=256$ variables, disjunctions of size $D=8$, and probability of mutexes $p= 0.121$ (higher plot means better performance). }
    \label{expr-mut-net-A}
\end{figure}

In the SAT solving phase, various performance characteristics of the SAT solver were measured such as runtime. The SAT solver has been given the time limit of 8000 seconds (approx. 2 hours 13 minutes). Preliminary experiments indicate that the runtime of the clique clustering and AMO encoding phases is negligible, hence there is no time limit of these phases. \footnote{All runtime measurements were done a system with a Ryzen 7 3GHz CPU cores and 32GB RAM running under Ubuntu Linux 19.}.

\subsubsection{Random Mutex Formulae}

A {\em random mutex formula} is characterized by tree parameters: $N$, the number of variables, $D$, the size of disjunctive clauses, and $p$, the probability of a mutex. The formula is denoted $\text{mutex-net}(N,D,p)$. The formula is simply constructed by declaring $N$ Boolean variables. Then each of possible $\frac{N \cdot (N -1)}{2}$ mutexes is added with probability of $p$. Such a formula is trivially satisfiable by setting all variables to $\mathit{FALSE}$. To make the formula more interesting we divide the set variables into $\lceil\frac{N}{D}\rceil$ disjoint subsets consisting of $D$ variables (the last group may consist of fewer variables).

\begin{figure}[h]
    \centering
    \includegraphics[trim={2.0cm 17.0cm 3.0cm 2.5cm},clip,width=0.9\textwidth]{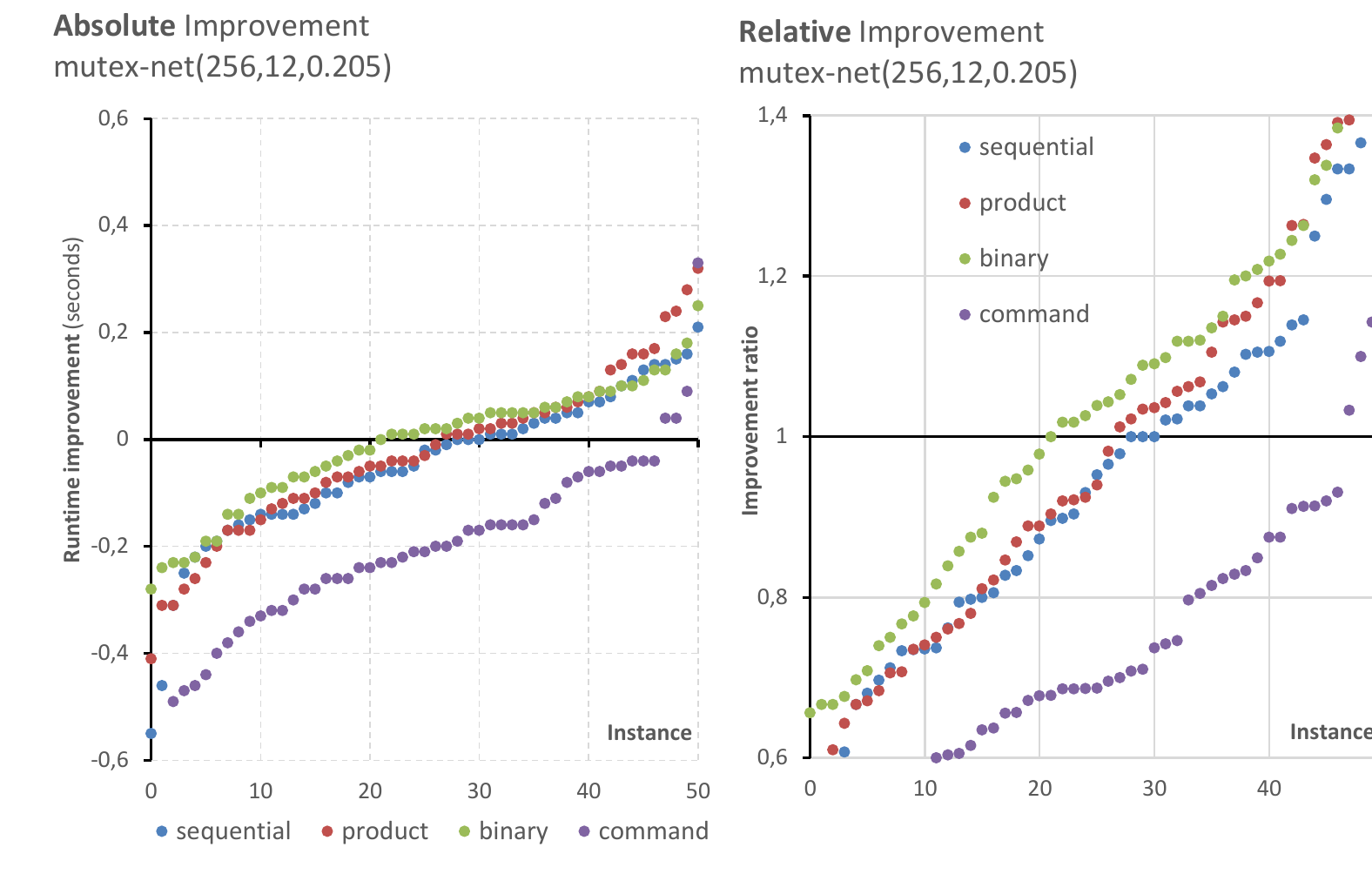}
    \caption{Absolute and relative improvement by AMOs using various encodings in random mutex network with $N=256$ variables, disjunctions of size $D=12$, and probability of mutexes $p = 0.205$.}
    \label{expr-mut-net-B}
\end{figure}

Mutex formulae tend to be relatively easy except setting of $p$ in certain narrow analogy of phases transition region. Runtime comparison for different encodings of detected AMO constraints on mutex formulae are presented in Figures \ref{expr-mut-net-A} and \ref{expr-mut-net-B} showing results for $\text{mutex-net}(256,8,0.121)$ and $\text{mutex-net}(256,12,0.205)$ respectively.

The probability of mutexes $p$ is selected to belong to the phase transition region; 60 random instances were generated for each parameter setting. The detected AMOs are encoded using all discussed encodings and compared to the baseline representation using the pair-wise encoding.

The sorted absolute and relative differences from the runtime of the base-line pair-wise encoding is shown. Generally, we cannot say there is universal improvement across all tested instances. Often the performance worsens with using the AMO constraints. The improvement is however in some cases up to $50\%$ compared to the runtime of the pair-wise encoding. We can also observe that using the commander and sequential encodings often results in worsening of performance. On the other hand the binary encoding and the product encoding tend to improve the situation.

Comparing the two classes of random mutex formulae we can observe that AMO substitution yields better results on  $\text{mutex-net}(256,8,0.121)$ while on $\text{mutex-net}(256,$ $12,0.205)$ worse performance after AMO substitution occur more frequently.

It is important to note that random mutex formulae are not especially suitable for finding large cliques. We used these instances to test the potential of AMO substitution under not very promising circumstances. The size of cliques identified in this experiments is usually 3 or 4 rarely 5. Still such small discovered cliques are shown to have potential for the AMO substitution.

\subsection{Clique Detection in Random Mutex Networks}

We also evaluated the performance of the relaxed clique detection separately from the SAT solving process. We focused on the size of cliques detected by the algorithm in this test. Random mutex networks $\text{mutex-net}(N,D,p)$ as defined in the previous section were used. The difference from the previous tests is that we take into account only the mutex clauses from $\text{mutex-net}(N,D,p)$ while larger disjunctions (those of size $D$) are ignored.

\begin{figure}[h!]
    \centering
    \includegraphics[trim={2.5cm 19.0cm 2.5cm 2.5cm},clip,width=1.0\textwidth]{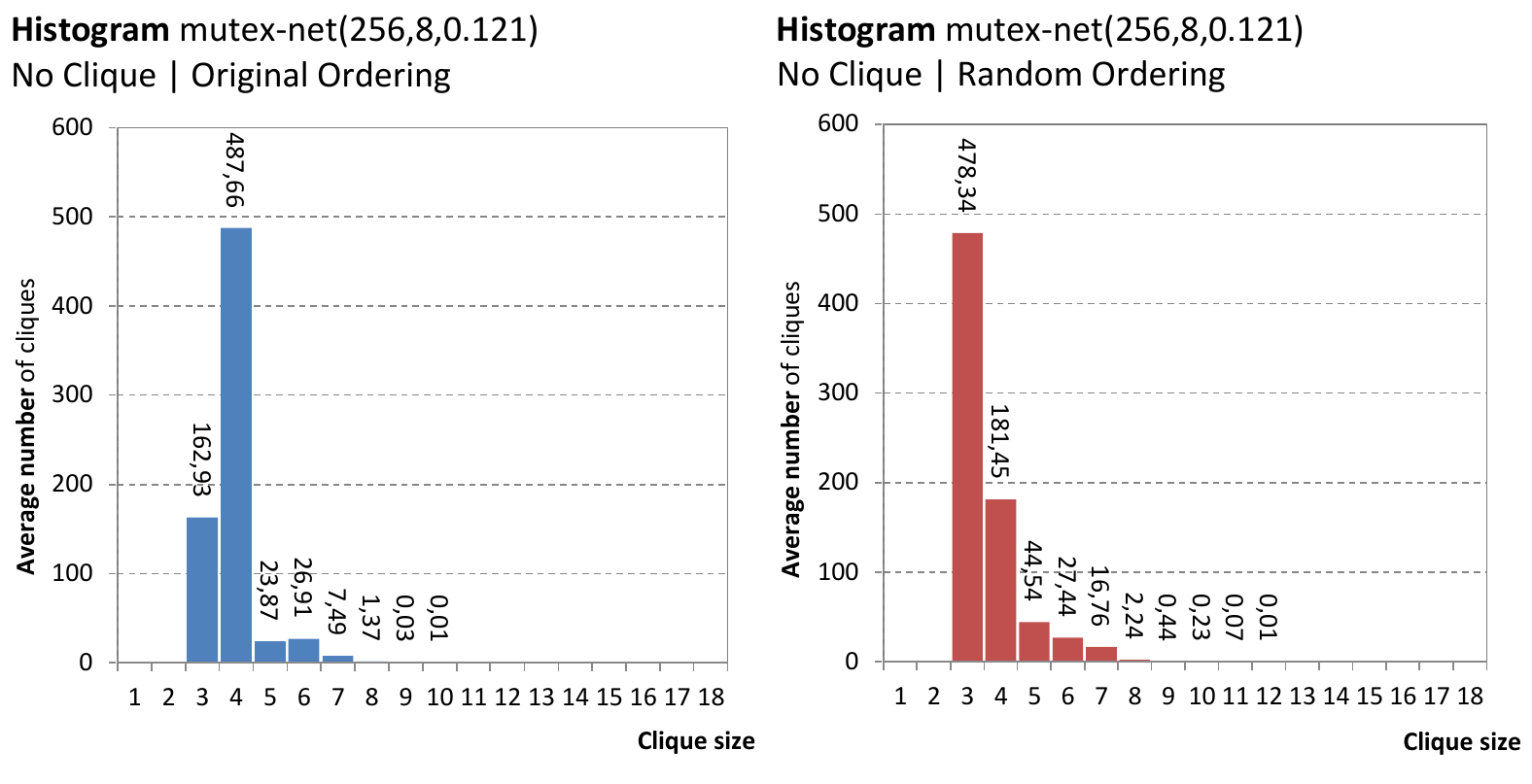}
    \caption{A histogram showing distribution of sizes of detected cliques in $\text{mutex-net}(256,8,0.121)$ where no cliques are explicitely introduced.}
    \label{expr-hist-noclique-A}
\end{figure}

\begin{figure}[h!]
    \centering
    \includegraphics[trim={2.5cm 19.0cm 2.5cm 2.5cm},clip,width=1.0\textwidth]{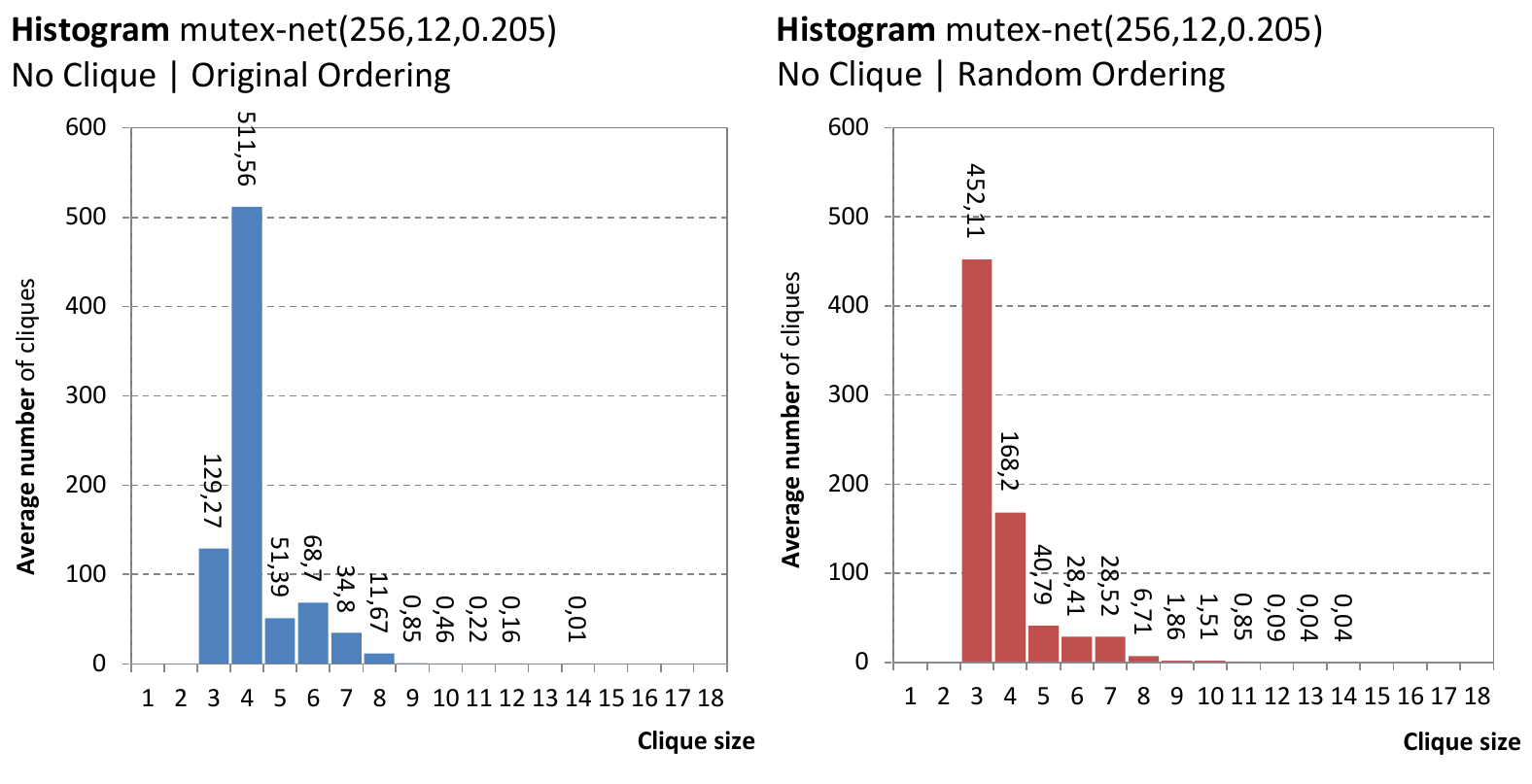}
    \caption{A histogram showing distribution of sizes of detected cliques in $\text{mutex-net}(256,12,0.205)$ where no cliques are explicitly introduced.}
    \label{expr-hist-noclique-B}
\end{figure}

Since the \textsc{Relaxed-Cliques} algorithm is sensitive to the ordering in which new mutexes arrive into the mutex network, the test is divided in two cases. In the first case, mutexes arrive in the lexicographic ordering of pairs indices of mutex variables - we refer to this case as an {\em original ordering}. In the second case, mutexes are permuted randomly - we refer to this case as a {\em random ordering}. Results for $\text{mutex-net}(256,8,0.121)$ and $\text{mutex-net}(256, 12,0.205)$ whose parameters are selected to belong to the aforementioned phase transition are shown in Figures \ref{expr-hist-noclique-A} and \ref{expr-hist-noclique-B}. The figures shows histogram of clique sizes across 100 randomly generated  $\text{mutex-net}(N,D,p)$. Each of the figures shows results for the original (left part) and random ordering (right part) of the arrival of mutexes into the mutex network.

It can be observed in both types of mutex networks that most of cliques are relatively small of sizes 3 or 4. Rarely a larger clique can be discovered - cliques of sizes up to 7 or 8 can be discovered. Moreover, the results clearly indicate that ordering of arrival of mutexes has a significant impact on what cliques are eventually discovered. Most of discovered cliques are of size 4 if the original ordering is used while most of cliques is of size 3 in the case of random ordering which can be again observed in both $\text{mutex-net}(256,8,0.121)$ and $\text{mutex-net}(256,$ $12,0.205)$ . On the other hand random ordering provides opportunity for a larger clique to grow.

\begin{figure}[h]
    \centering
    \includegraphics[trim={2.5cm 19.0cm 2.5cm 2.5cm},clip,width=1.0\textwidth]{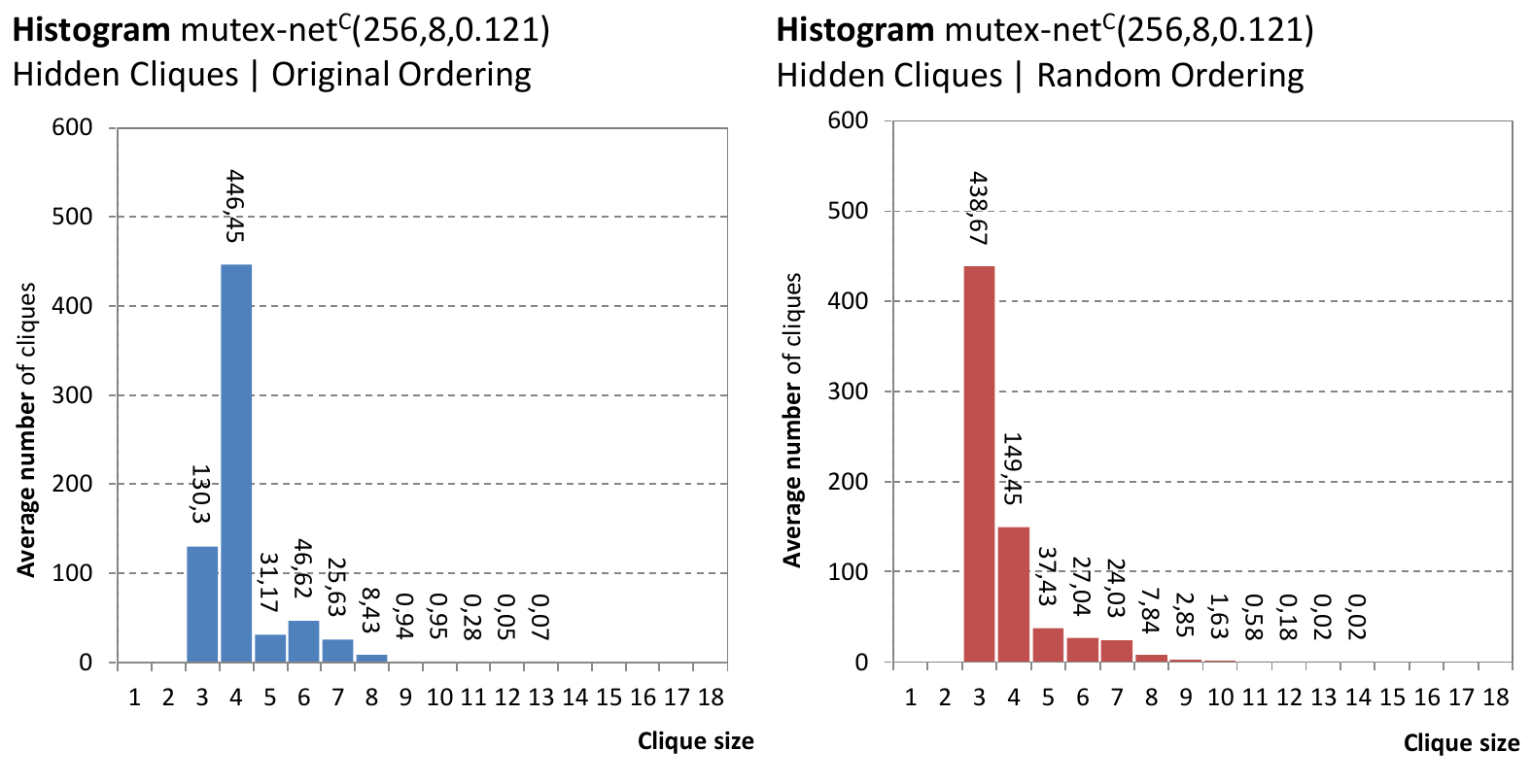}
    \caption{A histogram showing distribution of sizes of detected cliques in $\text{mutex-net}(256,8,0.121)$ where hidden cliques of size 8 are present.}
    \label{expr-hist-clique-A}
\end{figure}

The explanation for this behavior is that the original (lexicographic) ordering supports the growth of clique cluster around a variable that is shared across a sub-sequence of mutexes within the input mutex sequence. Such opportunity is less likely to occur when the ordering of mutexes in completely random. The explanation of detecting larger cliques with random ordering is that clusters in such case are more evenly distributed hence the chance of merging a pair large clusters is higher.

\begin{figure}[h]
    \centering
    \includegraphics[trim={2.5cm 19.0cm 2.5cm 2.5cm},clip,width=1.0\textwidth]{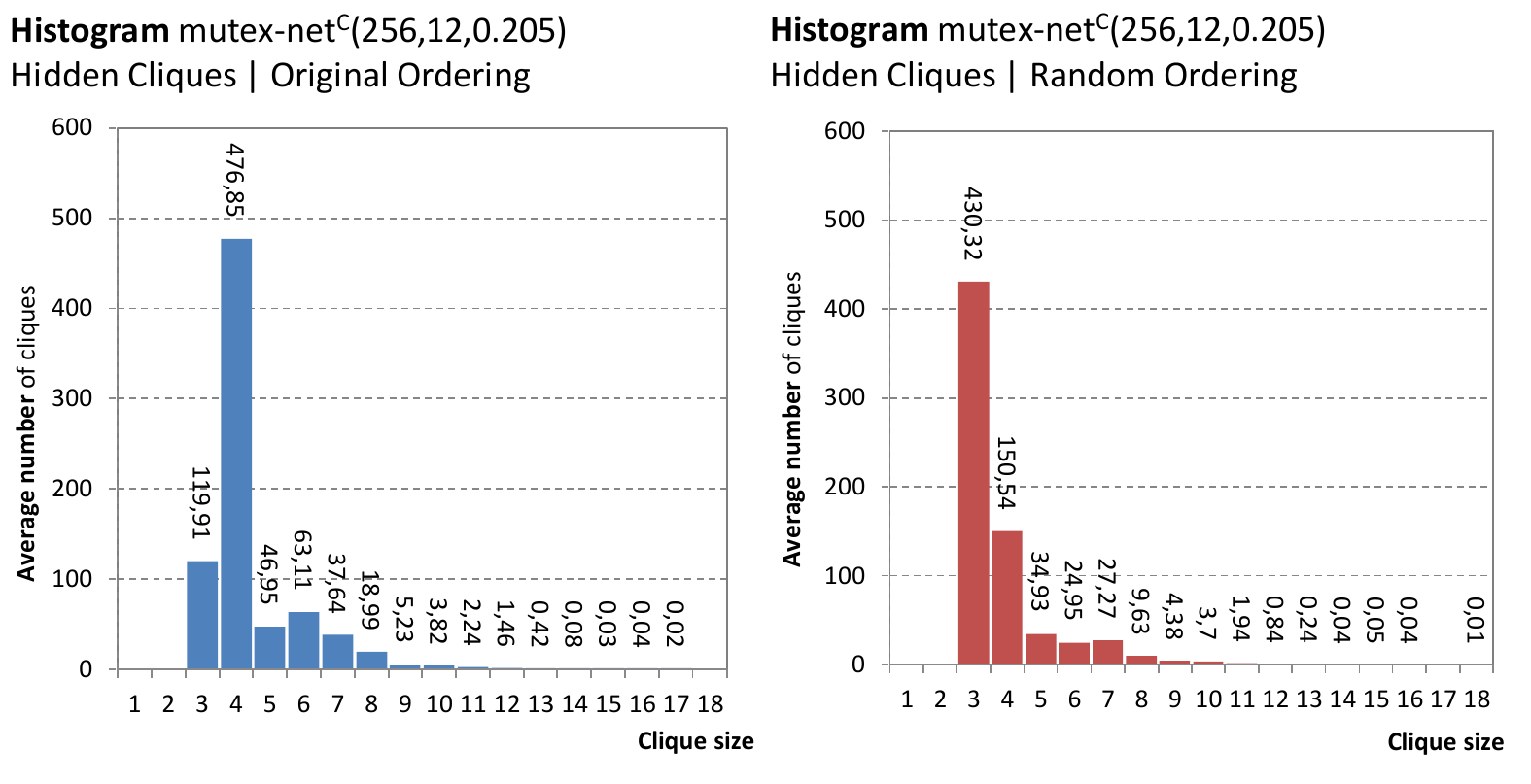}
    \caption{A histogram showing distribution of sizes of detected cliques in $\text{mutex-net}(256,12,0.205)$ where hidden cliques of size 12 are present.}
    \label{expr-hist-clique-B}
\end{figure}

The next test is focused on evaluation of discovering large cliques in random mutex networks. We use $\text{mutex-net}(N,D,p)$ as a basis but parameter $D$ is used for generating cliques.  Instead of introducing a disjunction of size $D$ a clique of mutexes of size $D$ is introduced. The clique can be regarded as {\bf hidden} in the mutex network. The resulting network will be denoted $\text{mutex-net}^{C}(N,D,p)$. Again we use the following setups: $\text{mutex-net}^{C}(256,8,0.121)$ and $\text{mutex-net}^{C}(256, 12,0.205)$ - the results are shown in Figures \ref{expr-hist-clique-A} and \ref{expr-hist-clique-B}. The original ordering corresponds first to adding mutexes randomly followed by adding cliques \footnote{Adding cliques as first would lead to their exact discovery as by example in Figure \ref{fig-cluster-A}. Adding random mutexes before leads to hiding the cliques in the network. }. The random ordering adds all mutexes from random phase and from cliques in a random order.

We can see in the results that clique of larger size can be detected in the networks compared to the case with no hidden cliques. However original cliques can be hardly recovered all. In $\text{mutex-net}^{C}(256,8,0.121)$ we can recover approximately 8 in 32 hidden cliques of size 8 when the original ordering is used. When the random ordering is used the chance is slightly lower. In $\text{mutex-net}^{C}(256, 12,0.205)$ only 1 clique of size 12 can be recovered from 20 such cliques hidden in the network.

The explanation for the observed behavior is that $\text{mutex-net}^{C}(256,8,0.121)$ is not as densely populated by random mutexes so there is still chance that the clique cluster grows around originally hidden cliques. This contrasts to the $\text{mutex-net}^{C}(256, 12,0.205)$ where two factors decrease chances to find the hidden cliques. First, these cliques are larger hence more successful steps are needed to detect them and second, the cliques are more overlaid by random mutexes. Hence harder to be detected by the relaxed clique algorithm. Despite not discovering all hidden cliques we cannot say the algorithm to be unsuccessful as it can find many smaller cliques and occasionally even larger ones.

\subsubsection{Classical Benchmarks}

\begin{table}[h]
    \centering
    \includegraphics[trim={1.0cm 17.0cm 1.0cm 1.5cm},clip,width=1.0\textwidth]{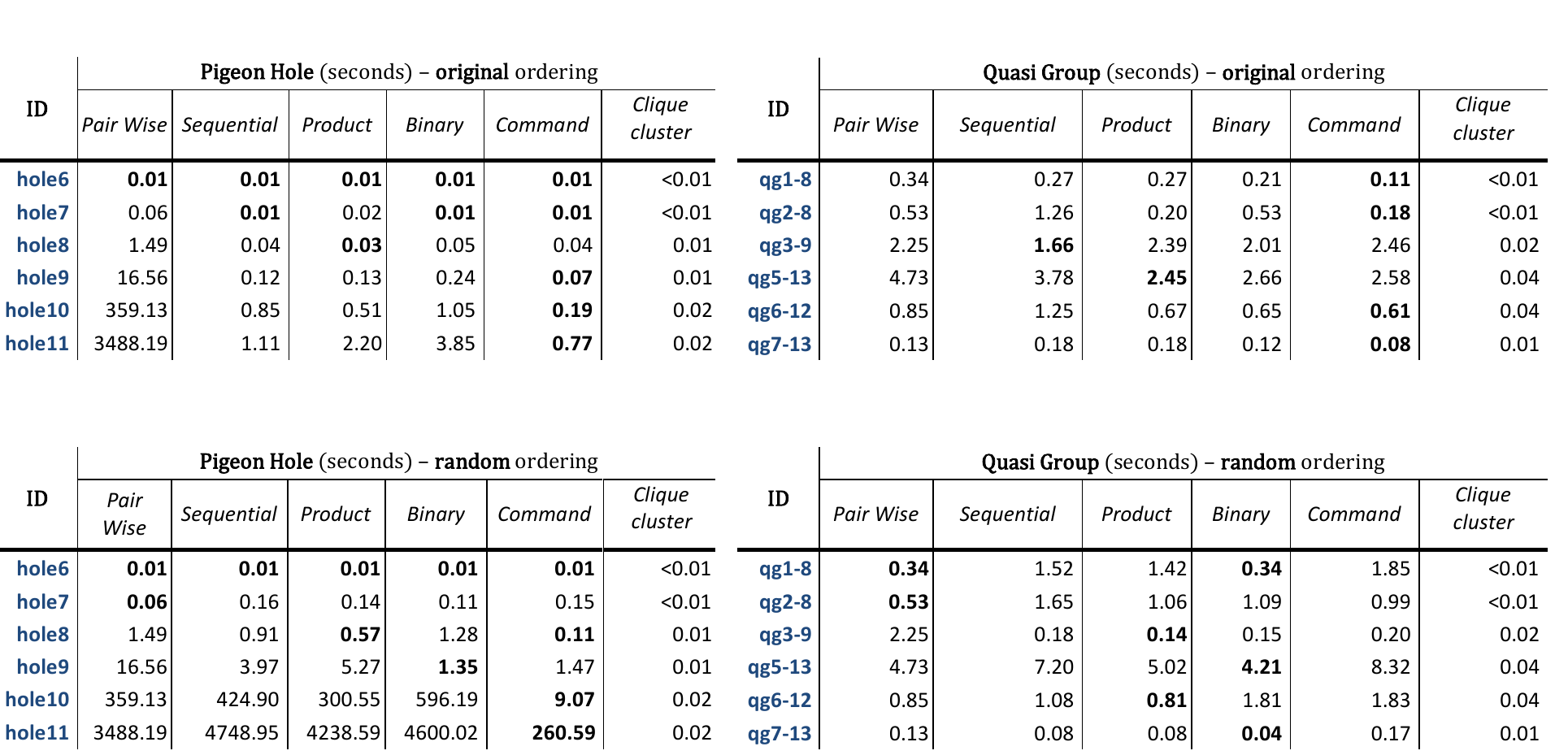}
    \caption{Runtime results for instances encoding the {\em Pigeon hole} priciple and {\em Quasi group} completion \cite{DBLP:conf/aaai/AnsoteguiVDFM04}.}
    \label{table-phole-qg}
\end{table}

The third test is focused on SAT solving of hard instances with relatively large mutex cliques hidden inside. The aim of this experiment is to verify if the \textsc{Relaxed-Cliques} algorithm is able to detect large enough mutex cliques so that their substituion by the AMO constraints results in a significant performance gain of the SAT solving phase.

We used standard benchmarks encoding the {\em pigeon hole} principle (denoted \texttt{hole}) where the question is whether $K+1$ pigeons can be placed in $K$ holes so that no two of them share a hole. This problem is known to be difficult for SAT solvers when the direct encoding is used \cite{DBLP:conf/sara/Surynek07}. In the direct encoding, there are variables $x_i^j$ encoding that $i$-th pigeon is placed in the $j$-th hole and a mutex network is introduced on top of these variables. Similarly various circuit routing problems are known to define difficult SAT instances containing cliques in their mutex networks \cite{DBLP:journals/tcad/AloulRMS03} (denoted \texttt{chnl} and \texttt{S3}). Circuit routing problems often encode sub-problems similar to the pigeon hole principle. Finally, {\em Quasi Group} (\texttt{qg}) instances encode construction of Latin squares \cite{DBLP:conf/aaai/AnsoteguiVDFM04}.

\begin{table}[t]
    \centering
    \includegraphics[trim={0.5cm 17.0cm 2.0cm 1.5cm},clip,width=1.0\textwidth]{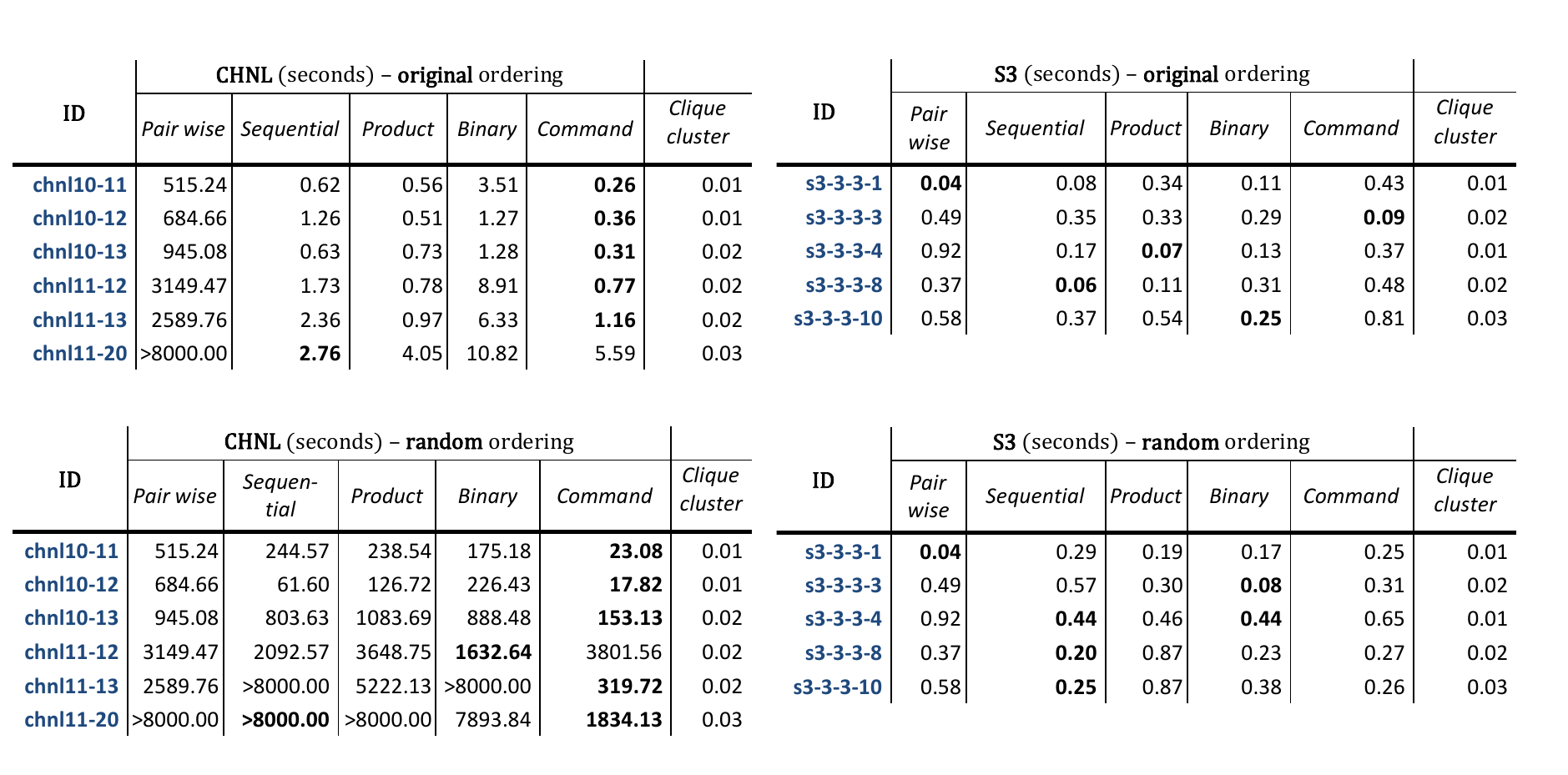}
    \caption{Runtime results for instances encoding the {\em Channel routing} and the {\em Global routing} instnaces \cite{DBLP:journals/tcad/AloulRMS03}.}
    \label{table-chnl-s3}
\end{table}

Runtime results are presented in Tables \ref{table-phole-qg} and \ref{table-chnl-s3}. We test the original ordering of mutexes and the random ordering. The former directly corresponds to the ordering of clauses in the input instance while the latter takes random permutation of mutex clauses in the input instance and performs relaxed clique detection with respect to this random permutation.

It can be observed for the original ordering that significant performance improvement is achieved for \texttt{hole} and \texttt{chnl} instances where the base-line pair-wise encoding often exceeds runtime of 1000 seconds while all encodings of the AMO constraint lead to runtimes in terms of seconds. The most significant improvement is achieved by using the commander encoding. This is quite surprising as in random mutex formulae the commander encoding has the worst performance. We need however to take into account that the size of cliques in \texttt{hole} and \texttt{chnl} instances is much larger (more than 10 variables) than in random mutex networks.

In \texttt{qg} and \texttt{S3} instances, the performance gain with the original ordering is less significant however using AMO substitution generally leads to better performance. The best runtime is almost in all cases achieved by some AMO encoding other than the pair-wise. If we compare individual encodings then the commander encoding and the binary encoding seem to provide the most consistent results. This is more in line with observations for random mutex networks where the binary encoding performs as best.

The results for random ordering of mutexes show generally worse performance of AMO substitution. The random ordering often leads to finding other cliques than those originally encoded in the input instance. Although the original clique covering is not detected, still relatively large cliques can be discovered. The performance gain, despite being less impressive than for the original ordering, can still be worthwhile.

We attribute the relatively good performance of AMO substitution in these classical benchmarks to two factors. First, the instances are difficult and hence there is room to improve the runtime (it is usually not good to use AMO substitution in quickly solvable instances as in such cases the overhead of clique clustering could play a role). Second, cliques in these instances are relatively large which gives chance the AMO encodings to significantly differ from the pair-wise encoding.

\section{Related Work}
Currently there seems to be a gap between works dealing with encodings of cardinality constraints and their automated detection. The notable exception is \cite{DBLP:conf/sat/BiereBLM14} where methods for automated rediscovering previously encoded AMOs using different encodings is presented. The difference from our work is that we are trying to detect AMOs in on-line mode and do not assume explicit presence of the AMO constraints in the encoding - even partial presence is valuable.

%\cite{DBLP:conf/sat/KuceraSV17}
Various works deal with efficient encodings of cardinality constraints and specially at-most-one constraints \cite{DBLP:conf/cp/SilvaL07,DBLP:conf/cp/BailleuxB03,DBLP:conf/cp/Sinz05}. The common effort is to find compact encoding (small size) that provides good support of unit propagation. As mentioned in \cite{DBLP:books/sp/Petke15} good propagation is often supported in direct encodings while small size is supported by logarithmic (log-space) encodings. Both factors are represented in our selection of AMO encodings.

Special focus on different encodings of the AMO constraint is given in \cite{DBLP:conf/soict/NguyenM15}. More encodings such as the {\em ladder encoding} and the {\em bimander encoding} are discussed and evaluated in this work. The difference from our work is that AMO constraints are identified in the mutex network manually. Automated detection of cliques in mutex network is done in \cite{DBLP:conf/sara/Surynek07} where a greedy algorithm is presented. The limitation of the greedy algorithm is that it is applicable in unsatisfiable case only where it can detect unsatisfiability by counting arguments without solving the formula. In the satisfiable case however, the method is not able to infer any new information.

\section{Conclusion}
We presented a method for on-line automated detection of At-Most-One constraints in mutex networks. Our AMO substitution method consists of a clique detection algorithm that is based on growing clusters that represent cliques. Any time a new mutex arrives, clique clusters are attempted to merge together to form a larger cluster, that is, a larger clique. The second major part of the AMO substitution method is encoding of detected cliques in mutex network as AMO constraints using one of the existent encodings.

We implemented the proposed method and performed experimental evaluation which indicates that even in random mutex networks containing small cliques, using more advanced encodings of the AMO constraint for automatically detected cliques has a potential to improve solving runtime. In hard instances containing large mutex cliques the method brings significant improvement in orders of magnitude. Moreover the clique detection and AMO encoding has negligible overhead according to our tests. Additional tests show that our method is able to find relatively large cliques even if the ordering of arriving mutexes is completely random (that is, mutexes forming a signle clique do not arrive together).

Future work include investigation of generalized mutex networks in which not only mutual exclusion between Boolean variables is considered but also mutual exclusion between literals. Hence any binary clause in such view will be treated as a mutex and included in the mutex network. While at the level of clique detection and AMO encoding the approach will not differ significantly. Different performance results may be expected. 

\section*{Acknowledgement}
\noindent 
This research has been supported by GA\v{C}R - the Czech Science Foundation, grant registration number 19-17966S. %We would like to thank anonymous reviewers for their valuable comments.

\bibliographystyle{splncs04}
\bibliography{bibfile}

\end{document}